\PassOptionsToPackage{unicode}{hyperref}
\PassOptionsToPackage{hyphens}{url}
\documentclass[
  11pt,
]{article}
\usepackage{amsmath,amssymb}
\usepackage{iftex}
\ifPDFTeX
  \usepackage[T1]{fontenc}
  \usepackage[utf8]{inputenc}
  \usepackage{textcomp} 
\else 
  \usepackage{unicode-math} 
  \defaultfontfeatures{Scale=MatchLowercase}
  \defaultfontfeatures[\rmfamily]{Ligatures=TeX,Scale=1}
\fi
\IfFileExists{lmodern.sty}{\usepackage{lmodern}}{}
\ifPDFTeX\else
\fi
\IfFileExists{upquote.sty}{\usepackage{upquote}}{}
\IfFileExists{microtype.sty}{
  \usepackage[]{microtype}
  \UseMicrotypeSet[protrusion]{basicmath}
  \microtypesetup{expansion=false}
}{}
\makeatletter
\@ifundefined{KOMAClassName}{
  \IfFileExists{parskip.sty}{%
    \usepackage{parskip}
  }{
    \setlength{\parindent}{0pt}
    \setlength{\parskip}{6pt plus 2pt minus 1pt}}
}{
  \KOMAoptions{parskip=half}}
\makeatother
\usepackage{xcolor}
\usepackage[margin=1in]{geometry}
\usepackage{longtable,booktabs,array}
\usepackage{calc} 
\usepackage{etoolbox}
\makeatletter
\patchcmd\longtable{\par}{\if@noskipsec\mbox{}\fi\par}{}{}
\makeatother
\IfFileExists{footnotehyper.sty}{\usepackage{footnotehyper}}{\usepackage{footnote}}
\makesavenoteenv{longtable}
\usepackage{stmaryrd}
\usepackage{booktabs}
\usepackage{graphicx}
\ifLuaTeX
  \usepackage{selnolig}  
\fi
\IfFileExists{bookmark.sty}{\usepackage{bookmark}}{\usepackage{hyperref}}
\IfFileExists{xurl.sty}{\usepackage{xurl}}{} 
\hypersetup{
  pdftitle={Intercepting the Kangaroo: Experimental Astrolinguistics with Constructed Lexicons, Active Probing, and Large Language Models as Informants and Hypothesis Proposers},
  pdfauthor={Francesco Cordella; Mauro Cappelli},
  hidelinks,
  pdfcreator={LaTeX via pandoc}}

\title{Intercepting the Kangaroo: Experimental Astrolinguistics with
Constructed Lexicons, Active Probing, and Large Language Models as
Informants and Hypothesis Proposers}
\author{Francesco
Cordella\thanks{ENEA, Centro Ricerche di Frascati, Frascati (RM), Italy. Corresponding author: francesco.cordella@enea.it. ORCID: 0000-0003-0731-5622.} \and Mauro
Cappelli\thanks{ENEA, Centro Ricerche di Frascati, Frascati (RM), Italy. mauro.cappelli@enea.it. ORCID: 0000-0002-4344-6067.}}
\date{}

\begin{document}
\maketitle
\begin{abstract}
Astrolinguistics --- communication with minds that categorize reality
differently from ours --- has been purely speculative since
Freudenthal's \emph{Lincos} (1960). We make it experimental. Two
language models with deliberately incompatible constructed lexicons (one
encoding shape, colour, and motion; the other fusing colour with motion,
encoding parity, and lacking shape) serve as informants with complete
ground truth, while a fully scripted orchestrator translates between the
two category systems. The central failure mode is the \emph{kangaroo
effect}: the silent attachment of a word to the wrong referent ---
Quine's indeterminacy of translation, operationalized. Across 400+
simulated and live runs, a protocol combining cross-situational
hypothesis elimination, pre-registered predictive probes, active scene
selection, a stricter recovery round, and quarantine produced no
undetected mistranslations under the tested conditions, while exceeding
the coverage of a passive baseline (d = 0.62). Deliberately injected
kangaroo traps defeated naive ostension and pure statistical learning in
100\% of runs, while the full protocol intercepted every decoy (mean
detection: exchange 40.6) and, where discriminating evidence is
ontologically unavailable, declared Quinean equivalence classes instead
of guessing. Under informant noise the result degrades gracefully: zero
kangaroos persist up to 2\% per-word noise; at 10\%, the protocol
predominantly abstains rather than errs. Finally, words outside the
scripted hypothesis space (a history-dependent relational term and an
XOR contextual homonym) are recovered by a generate-and-test loop in
which an LLM proposes rules and the script verifies them: coverage
scales with proposer capability (0\% \(\rightarrow\) 18\%
\(\rightarrow\) 72\% \(\rightarrow\) 100\%) while undetected
mistranslations remained at zero throughout. In the tested conditions,
correctness is a property of the protocol; coverage is a property of the
instruments.

\end{abstract}

\hypertarget{introduction}{%
\subsection{1. Introduction}\label{introduction}}

When Freudenthal (1960) designed Lincos, a language for cosmic
intercourse, he proceeded from mathematics through physics toward
behaviour --- and stopped, by his own admission, where evaluative and
sociological concepts begin. Sixty-five years later the discipline he
founded remains in the condition he left it: a design exercise without
an experimental arm. No astrolinguistic protocol has ever been
\emph{tested}, for the obvious reason that no alien interlocutor has
ever been available.

This paper proposes and demonstrates a way around that obstacle. The
essential difficulty of communicating with a radically different mind is
not the absence of a shared channel but the absence of shared
\emph{categories}: the other mind may cut the world along joints we do
not have, fuse distinctions we keep separate, and lack dimensions we
consider basic. That difficulty can be manufactured in the laboratory.
We construct two artificial lexicons with deliberately incompatible
category systems, assign each to a large language model (LLM) acting as
a native informant, and give a fully scripted learner the task of
translating between them. Because the lexicons are constructed, the
experimenter holds complete ground truth --- every claimed translation
can be checked exactly --- which is precisely what fieldwork with a real
alien (or a real undeciphered language) can never offer.

The failure mode that organizes the entire enterprise is what we call
the \textbf{kangaroo effect}: the silent attachment of a word to the
wrong referent, named after the (apocryphal but instructive) story of
Cook's expedition asking for the name of the animal and recording the
answer without ever being able to verify what question the informant
thought he was answering. The kangaroo effect is our operational
rendering of Quine's (1960) indeterminacy of translation: two hypotheses
about a word's referent can agree on every scene observed so far and
diverge only on scenes never shown. Our design principle, validated
throughout this paper, is that \emph{the kangaroo cannot be prevented
--- it can only be intercepted}: no amount of passive observation rules
it out, but a protocol that registers predictions before testing them,
actively constructs the scenes on which rival hypotheses disagree, and
is willing to revoke its own promotions can drive the rate of
\emph{undetected} mistranslations to zero --- under the conditions we
test, and with a degradation profile that we measure rather than assume.

This paper delivers the behavioural layer --- \textbf{Level A} --- of a
two-level programme. The programme's target is alignment between the
latent representation spaces of different models under category mismatch
(\textbf{Level B}); the protocol presented here is designed to serve
there as the behavioural verification layer for claimed alignments,
which is why every mechanism in it is built to \emph{certify}, rather
than assume, that two systems are using the same concept. The
micro-world that hosts the experiments is deliberately minimal, and the
minimality is methodological rather than incidental: with 32 scenes and
complete ground truth, every promotion, every kangaroo, and every
declared underdetermination is exactly checkable --- the smallness is
what converts a philosophical worry into a measurement.

The contributions are fivefold. \textbf{First}, a complete experimental
protocol (Section 4) --- cross-situational hypothesis elimination,
contrastive minimal pairs, pre-registered predictive probes, active
scene selection, a stricter second ``recovery'' round, and a
quarantine/revocation mechanism --- implemented entirely in
deterministic script, with the LLMs confined to the role of informants.
\textbf{Second}, a four-arm campaign over 50 random worlds (Section 5.1)
showing that active probing drove the kangaroo rate to zero across all
50 tested worlds, that a recovery round raises coverage \emph{above} the
passive baseline at zero kangaroos, and that these two mechanisms are
synergistic rather than additive: the recovery round without active
selection recovers little and can even promote a false translation.
\textbf{Third}, deliberately injected kangaroo traps (Section 5.2),
which defeated naive ostension and pure statistical learning in 100\% of
runs while the full protocol intercepted every decoy in the tested runs,
and which let us operationalize Quinean underdetermination as a
measurable, \emph{declarable} outcome distinct from both success and
failure. \textbf{Fourth}, an ablation on words that lie outside the
scripted hypothesis space (Section 5.3): a history-dependent relational
word and an XOR-structured contextual homonym, unreachable by the base
learner, are recovered by a generate-and-test architecture in which an
LLM \emph{proposes} candidate rules in a restricted formal language and
the script \emph{verifies} them against the log and against fresh
pre-registered probes. Coverage on these hard words scales with the
capability of the proposer --- from 0\% with no proposer, through 18\%
(Claude Haiku 4.5) and 72\% (Claude Sonnet 4.6), to 100\% with a
hand-coded oracle space --- while the rate of undetected mistranslation
stayed at zero across the entire tested ladder. \textbf{Fifth}, a
robustness study under injected informant noise (Section 5.4) that
delimits the validity domain of the zero-kangaroo result honestly: the
guarantee survives intact up to 2\% per-word channel noise and degrades
thereafter by abstention far more than by error. The slogan that
summarizes the empirical content of this paper --- \emph{correctness is
a property of the protocol; coverage is a property of the instruments}
--- is therefore a claim about the tested regime, with measured
boundaries.

\hypertarget{related-work-and-positioning}{%
\subsection{2. Related work and
positioning}\label{related-work-and-positioning}}

\textbf{Cross-situational learning.} The learning mechanism at the core
of our orchestrator is not new: it is cross-situational learning (XSL)
by hypothesis elimination, given its first precise computational
formulation by Siskind (1996), who showed that intersecting candidate
meanings across situations, combined with covering and exclusivity
constraints, can solve child-scale lexical acquisition under worst-case
assumptions. Siskind's algorithm also anticipated, by its very
limitation, our central addition: in his framework a corrupted lexical
entry cannot be directly repaired --- recovery proceeds by spawning
fresh sense entries while the corrupted ones decay --- whereas our state
machine makes revocation a first-class, measurable transition
(quarantine). Fazly, Alishahi and Stevenson (2010) recast XSL
probabilistically and analysed Siskind's brittleness under noise; Smith,
Smith and Blythe (2011) confirmed XSL experimentally in human learners.
We cite this thirty-year tradition as foundation, not as contribution.

\textbf{Emergent communication and referential games.} A second adjacent
tradition lets artificial agents \emph{develop} communication protocols
in referential games (Lazaridou et al., 2016; Steels, 2015), recently
with LLM agents (Kouwenhoven et al., 2024). Our setting inverts the
direction of fit: nothing emerges --- both lexicons pre-exist and are
held fixed; the task is \emph{translation between} two given systems,
and the LLM is the informant whose interpretive behaviour is itself an
object of study, not a learner.

\textbf{Latent-space alignment.} The programme of which this paper is
Level A continues toward alignment of representation spaces (Level B),
for which the relevant tools are relative representations (Moschella et
al., 2023; Maiorca et al., 2023) and anchor-based cross-lingual mapping
(Conneau et al., 2017). Experimental semiotics (Galantucci, 2005) and
post-Lincos astrolinguistics (Ollongren, 2013) complete the map.

\textbf{Concept alignment and AI safety.} Although developed from an
astrolinguistic question, the kangaroo effect is an experimental
instance of a problem now central to AI alignment: how to verify that
two systems are using the \emph{same} concept, rather than concepts that
merely agree on the cases observed so far (Sucholutsky et al., 2023).
The Eliciting Latent Knowledge problem (Christiano et al., 2021) is the
same structure seen from inside a single model --- a reporter that
answers correctly on the training distribution may be tracking the wrong
referent --- and the generate-and-test architecture of Section 5.3, in
which an untrusted generator's proposals are accepted only after
surviving scripted falsification, shares its logic with
verification-based oversight schemes such as debate (Irving et al.,
2018). Ontology matching (Euzenat and Shvaiko, 2013) addresses category
mismatch between knowledge bases with structural and extensional
heuristics; our setting differs in having an interactive informant,
which makes active falsification possible, and complete ground truth,
which makes its value measurable.

\textbf{The claimable novelty} is therefore not the use of LLMs, nor the
learning mechanism, but the \emph{configuration}, which to our knowledge
has never been assembled: (i) an informant with deliberately constructed
\emph{transversal} categories (fused words, missing dimensions, parity
terms) and complete ground truth, shifting the question from ``how is a
language learned'' to ``how does one translate between different ways of
cutting the world''; (ii) the LLM as informant-that-interprets rather
than learner, making phenomena such as presupposition of plurality
experimental observables; (iii) the anti-kangaroo machinery itself as
the evaluated object, with kangaroo rate, detection time, and declared
underdetermination as metrics; (iv) the bridge XSL \(\leftrightarrow\)
emergent communication \(\leftrightarrow\) astrolinguistics, with
LLM-proposed hypotheses as the controlled extension point.

\hypertarget{the-micro-world-and-the-constructed-lexicons}{%
\subsection{3. The micro-world and the constructed
lexicons}\label{the-micro-world-and-the-constructed-lexicons}}

The world consists of 32 scenes, the Cartesian product of four
attributes: shape (triangle/circle), colour (red/blue), motion
(fast/still), and number (1--4), the last observed by the learner only
through its parity. Scenes are rendered to the informants as short
natural-language descriptions.

Two lexicons are constructed to be categorially incompatible (Table 1).
L1, the ``human'' lexicon (spoken by Claude Haiku 4.5), encodes shape,
colour, and motion as independent words and does not encode number. L2,
the ``alien'' lexicon (spoken by GPT-4o-mini), \emph{fuses} colour and
motion into four portmanteau words, encodes the parity of the number,
and has no word for shape at all. The learner's task is to reconstruct
the L2 dictionary --- each word's referent, expressed as a conjunction
of attribute--value atoms --- from interaction alone.

\textbf{Table 1.} The two constructed lexicons. L2's categories are
transversal to L1's: colour and motion are fused, parity is introduced,
shape is absent.

\begin{longtable}[]{@{}lll@{}}
\toprule\noalign{}
System & Word & True referent \\
\midrule\noalign{}
\endhead
\bottomrule\noalign{}
\endlastfoot
L1 (human) & KEPO / KIMU & shape = triangle / circle \\
L1 & KARI / KOLA & colour = red / blue \\
L1 & KESU / KANO & motion = fast / still \\
L2 (alien) & ZUMU & colour = red \(\wedge\) motion = fast \\
L2 & ZAKA & colour = red \(\wedge\) motion = still \\
L2 & ZIBO & colour = blue \(\wedge\) motion = fast \\
L2 & ZEFU & colour = blue \(\wedge\) motion = still \\
L2 & TAK / TIN & parity = even / odd \\
\end{longtable}

\textbf{Word-form selection.} The word forms are nonce strings chosen
under four criteria. First, they carry no meaning in the languages the
informants were trained on --- a non-trivial requirement when the
informant is an LLM, since a form resembling a real morpheme would
import prior associations into emission behaviour. Perfect neutrality is
unattainable for a model trained on human text (TAK, for instance, is
Polish for ``yes'\,'), so neutrality is verified empirically rather than
assumed: any residual association would surface as an execution-fidelity
deficit, and measured fidelity is 100\%. Second, the family initials (K-
for L1, Z- for the L2 fusions, T- for parity) are a log-readability
convenience for the experimenter and are invisible to the learner, which
treats words as opaque atomic symbols in the tradition of Siskind
(1996), so no property of the word surface can either help or mislead
it. Third, within each family every pair of forms differs by edit
distance \(\ge\) 2, so a single-character corruption cannot turn one
valid word into another valid word. Fourth, short CVCV shapes keep
utterances compact and tokenizer-friendly, in line with nonce-word
practice in experimental semiotics and human cross-situational learning
studies (Smith, Smith and Blythe, 2011).

Formally, let \(A\) be the set of attribute--value atoms. The base
hypothesis space contains all conjunctions of at most two atoms over
distinct attributes:

\[\mathcal{H}_{0}=\bigl\{\,h\subseteq A \;:\; 1\le\lvert h\rvert\le 2,\ \ (a,v),(a',v')\in h \Rightarrow a\neq a'\,\bigr\},\qquad \lvert\mathcal{H}_{0}\rvert = 32. \qquad (1)\]

All informant calls use temperature 0 --- the decoding setting at which
the model always selects its single most probable next token, making its
output effectively deterministic --- and a fixed system prompt
containing the lexicon. The setting is methodological rather than
incidental: the informant must behave as a rule applier, not as a
stochastic generator, so that runs are reproducible and any variability
in the channel is the variability we deliberately inject and measure
(Section 5.4), never uncontrolled sampling noise. A rule-based simulated
speaker (SIM) that applies the lexicon perfectly serves as the control
arm throughout. \emph{Execution fidelity} --- agreement between each
emitted utterance and the lexicon's rule --- is computed automatically
from the logs and separates speaker errors from translation errors. In
all experiments reported here, fidelity is 100\% (after a one-line
prompt repair of GPT-4o-mini's presupposition of plurality, documented
in earlier work), so that every residual error is attributable to the
protocol, not to the informant.

\hypertarget{the-anti-kangaroo-protocol}{%
\subsection{4. The anti-kangaroo
protocol}\label{the-anti-kangaroo-protocol}}

Figure 1 shows the protocol as a state machine over \emph{anchors}
(word--hypothesis-set pairs). Each newly heard word \(w\) enters as
PROVISIONAL with \(\mathcal{H}_{w}=\mathcal{H}_{0}\).

\begin{figure}[t]\centering
\includegraphics[width=0.95\linewidth]{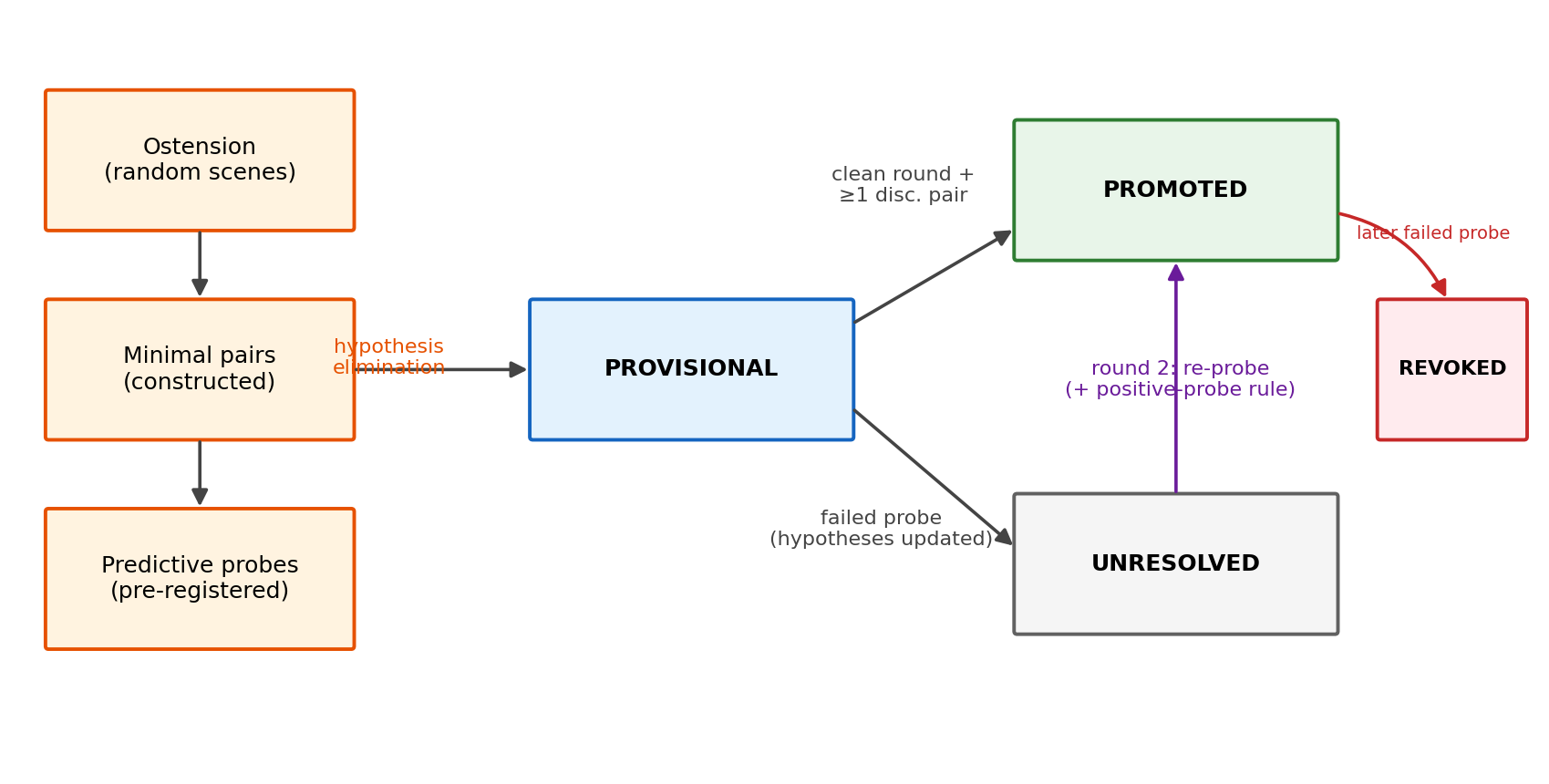}
\caption{The anti-kangaroo state machine. Evidence flows from ostension, constructed minimal pairs, and pre-registered probes; promotion requires a clean probe round plus contrastive evidence; failures send the anchor to UNRESOLVED with corrected hypotheses; the recovery round (purple) re-probes unresolved anchors under a stricter criterion; a promoted anchor that later fails a probe is quarantined (REVOKED).}\label{fig:1}
\end{figure}

\textbf{Elimination.} Whenever scene \(s\) is shown and the informant's
utterance \(u(s)\) is observed, every anchor is updated by
cross-situational elimination --- a hypothesis survives iff its truth on
\(s\) matches the word's presence in \(u(s)\):

\[\mathcal{H}_{w}\;\leftarrow\;\bigl\{\,h\in\mathcal{H}_{w}\;:\;\llbracket h(s)\rrbracket=\llbracket w\in u(s)\rrbracket\,\bigr\}. \qquad (2)\]

With a perfectly faithful speaker, (2) can never eliminate the true
referent; kangaroos arise only from the \emph{choice} among surviving
hypotheses.

\textbf{Minimal pairs.} For each attribute occurring in surviving
hypotheses, the orchestrator constructs a scene pair identical except
for that attribute; a word that co-varies with the change yields a
\emph{discriminating pair}, a precondition for promotion.

\textbf{Pre-registered predictive probes.} Before each probe scene \(s\)
is shown, the learner registers the prediction

\[\hat{p}_{w}(s)\;=\;\bigvee_{h\in\mathcal{H}_{w}} h(s), \qquad (3)\]

then queries the informant and compares. A failed probe both blocks
promotion in the current round and \emph{informs}: the offending scene
is fed back through (2), so the hypothesis set exits the failure already
corrected.

\textbf{Active scene selection.} In the active condition, the probe
scene is chosen to maximize, lexicographically, (i) the \emph{split} of
the surviving hypotheses, (ii) expected presence of the word, and (iii)
novelty of unconstrained attributes:

\[s^{*}=\arg\max_{s}\ \left(\min(V_{s},\,\lvert\mathcal{H}_{w}\rvert-V_{s}),\ \ \mathbb{1}[V_{s}>0],\ \ \mathrm{nov}(s)\right),\qquad V_{s}=\sum_{h\in\mathcal{H}_{w}}\llbracket h(s)\rrbracket . \qquad (4)\]

The split criterion is the anti-kangaroo mechanism in its purest form:
it sends the conversation precisely to the scenes on which rival
hypotheses \emph{disagree} --- the scenes a kangaroo needs the learner
never to see.

\textbf{Promotion and the recovery round.} In probing round \(r\), with
\(M\) probes per anchor, promotion requires a clean round plus
contrastive evidence; from the second round on, a
\emph{positive-evidence} condition is added so that an accidentally
consistent hypothesis cannot be promoted on absences alone:

\[\begin{aligned}\text{PROMOTED at round } r \iff\ &\#\{\text{correct}\}=M \;\wedge\; \#\{\text{failed}\}=0 \;\wedge\; \#\{\text{disc. pairs}\}\ge 1\\ &\wedge\; \left(r\ge 2 \Rightarrow \#\{\text{positive probes}\}\ge 1\right). \qquad (5)\end{aligned}\]

Round 2 re-probes only non-promoted anchors, whose hypothesis sets have
already been corrected by their round-1 failures. A promoted anchor that
later fails any probe is REVOKED (quarantine); in no run reported in
this paper was a false translation ever left promoted.

\textbf{Outcome classification.} Let \(r_{w}\) be the true referent. The
class-aware classification, introduced in Section 5.2 and used wherever
underdetermination is possible, is:

\[\mathrm{outcome}(w)=\begin{cases}\text{unresolved} & \text{if not promoted}\\[2pt] \text{KANGAROO} & \text{if promoted and } r_{w}\notin\mathcal{H}_{w}\\[2pt] \text{CORRECT} & \text{if promoted and } \mathcal{H}_{w}=\{r_{w}\}\\[2pt] \text{UNDERDETERMINED} & \text{if promoted, } r_{w}\in\mathcal{H}_{w},\ \lvert\mathcal{H}_{w}\rvert>1.\end{cases} \qquad (6)\]

A word is a kangaroo only if the true referent is \emph{provably
excluded}; if it survives alongside extensionally indistinguishable
rivals, the honest output is the equivalence class --- Quinean
underdetermination made operational.

\textbf{Determinism and reproducibility.} Two engineering lessons proved
methodologically substantive. First, Python's hash randomization changes
set-iteration order between processes; every iteration over sets and
dictionaries is therefore canonically sorted, including --- a latent bug
found and fixed during this work --- the Occam tie-break in the referent
estimate, which previously depended on \texttt{PYTHONHASHSEED}. Second,
the random baseline is a lottery sensitive to the RNG stream:
comparisons are valid only at parity of code version. Section 5.3
extends this rule to prompts and parsers (``extended protocol parity'').
With these conventions, every SIM result in this paper reproduces to the
last decimal across machines, and the LLM arm reproduces the SIM arm
character-for-character at 100\% fidelity.

\hypertarget{experiments}{%
\subsection{5. Experiments}\label{experiments}}

All campaigns use seeds \(0,\dots,N-1\); SIM campaigns are free and
fully deterministic; LLM campaigns cost cents to a few euros in API
calls (the entire Level A programme, including all pilots, cost under
EUR 5).

\hypertarget{the-recovery-round-and-the-four-arm-campaign}{%
\subsubsection{5.1 The recovery round and the four-arm
campaign}\label{the-recovery-round-and-the-four-arm-campaign}}

The trade-off discovered in earlier 10-seed pilots was that active
selection eliminated kangaroos but lost some coverage: probe failures
correct the hypotheses yet block promotion, leaving anchors ``unresolved
with the right referent''. The recovery round (Eq. 5, \(r=2\)) is
designed to convert exactly these cases. Table 2 and Figure 2 report the
full \(2\times 2\) campaign --- probe selection (random/active)
\(\times\) recovery round (off/on) --- at 50 seeds per arm.

\textbf{Table 2.} Four-arm campaign, SIM speaker, 50 seeds per arm (mean
± SD per run; 6 L2 words per run). R2 = recoveries at round 2.

\begin{longtable}[]{@{}
  >{\raggedright\arraybackslash}p{(\columnwidth - 10\tabcolsep) * \real{0.1667}}
  >{\raggedright\arraybackslash}p{(\columnwidth - 10\tabcolsep) * \real{0.1667}}
  >{\raggedright\arraybackslash}p{(\columnwidth - 10\tabcolsep) * \real{0.1667}}
  >{\raggedright\arraybackslash}p{(\columnwidth - 10\tabcolsep) * \real{0.1667}}
  >{\raggedright\arraybackslash}p{(\columnwidth - 10\tabcolsep) * \real{0.1667}}
  >{\raggedright\arraybackslash}p{(\columnwidth - 10\tabcolsep) * \real{0.1667}}@{}}
\toprule\noalign{}
\begin{minipage}[b]{\linewidth}\raggedright
Arm
\end{minipage} & \begin{minipage}[b]{\linewidth}\raggedright
Correct
\end{minipage} & \begin{minipage}[b]{\linewidth}\raggedright
Kangaroos
\end{minipage} & \begin{minipage}[b]{\linewidth}\raggedright
Unresolved
\end{minipage} & \begin{minipage}[b]{\linewidth}\raggedright
Exchanges
\end{minipage} & \begin{minipage}[b]{\linewidth}\raggedright
R2 (false)
\end{minipage} \\
\midrule\noalign{}
\endhead
\bottomrule\noalign{}
\endlastfoot
Random & 4.18 ± 0.92 & 0.24 ± 0.43 & 1.58 ± 0.84 & 53.6 ± 5.2 & --- \\
Random + R2 & 4.36 ± 0.94 & 0.26 ± 0.49 & 1.38 ± 0.85 & 58.4 ± 4.9 & 10
(1) \\
Active & 4.06 ± 0.93 & \textbf{0.00 ± 0.00} & 1.94 ± 0.93 & 53.6 ± 5.2 &
--- \\
\textbf{Active + R2} & \textbf{4.72 ± 0.81} & \textbf{0.00 ± 0.00} &
1.28 ± 0.81 & 59.5 ± 4.9 & 33 (0) \\
\end{longtable}

\begin{figure}[t]\centering
\includegraphics[width=0.75\linewidth]{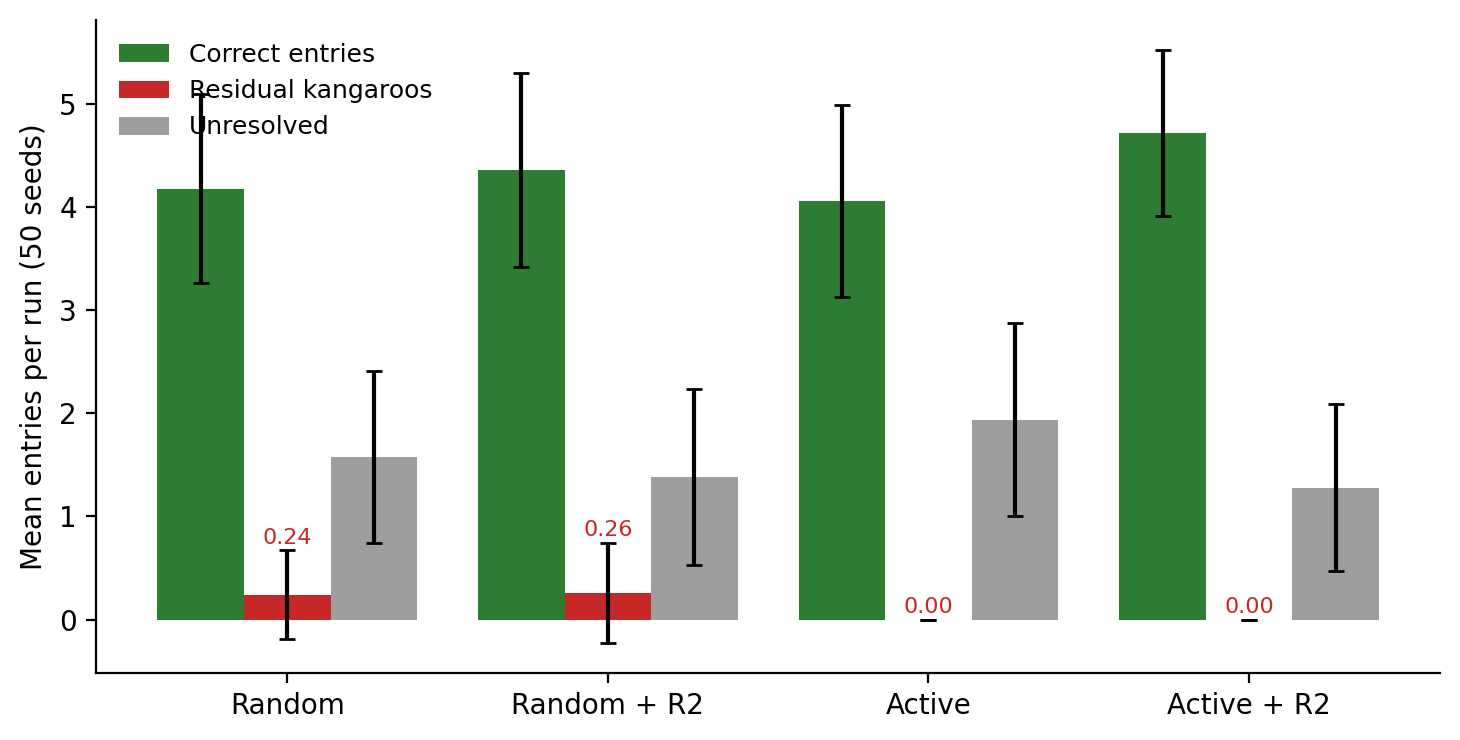}
\caption{Outcome composition per arm (mean ± SD over 50 seeds). Active probing yielded zero kangaroos over 50 seeds; the recovery round restores --- and surpasses --- the coverage of the random baseline at zero kangaroos, for $\approx$ 11\% more exchanges.}\label{fig:2}
\end{figure}

Three findings. \textbf{(i) Recovery exceeds the target.} Active + R2
reaches 4.72 correct entries per run, above the random baseline's 4.18
(\(\Delta\) = +0.54, 95\% bootstrap CI {[}0.20, 0.86{]}, Cohen's d =
0.62; vs the active arm alone: \(\Delta\) = +0.66, CI {[}0.32, 1.00{]},
d = 0.75), with zero kangaroos in 50 runs and all 33 round-2 promotions
correct, at a cost of \(\approx\) 6 extra exchanges per run. The
kangaroo contrast is itself significant: 12/50 random-arm runs contain
at least one kangaroo versus 0/50 in Active + R2 (Fisher exact p = 2.3
\(\times\) 10\(^{-4}\)), and the observed zero over 300 word-runs bounds
the per-word-run kangaroo probability below 1.0\% (rule of three, 95\%
confidence). \textbf{(ii) The mechanisms are synergistic, not additive.}
In the random arm, R2 recovers only 10 entries --- and one of them is a
\emph{kangaroo}: a round-2 promotion of ZUMU with estimated referent red
\(\wedge\) triangle instead of red \(\wedge\) rapid (seed 38). The probe
log shows why: of the three random round-2 probes, the single positive
probe fell on a scene (a fast red triangle) on which \emph{both}
surviving hypotheses predict presence, so the positive-evidence
condition of (5) was satisfied by a non-discriminating positive. Active
selection cannot commit this error by construction, because its primary
criterion (4) targets maximum disagreement. The safety of the recovery
round is therefore a property of the \emph{pair} (recovery, active
selection) --- an argument that is theoretical as well as empirical.
\textbf{(iii) Protocol, not informant.} The live campaign (GPT-4o-mini
as L2, active + R2, 10 seeds, 584 calls, 100\% fidelity) reproduces the
SIM campaign character-for-character: same per-seed outcomes, same
recoveries, same exchange counts. The kangaroo phenomenon, and its
remedy, belong to the protocol, not to the species of the informant ---
extending to the recovery mechanism the equivalence previously
established for the base protocol.

\hypertarget{injected-kangaroo-traps}{%
\subsubsection{5.2 Injected kangaroo
traps}\label{injected-kangaroo-traps}}

The seed-38 kangaroo arose naturally; Experiment 3 injects it
deliberately. A \emph{trap} is a pair (word, decoy): an alternative
hypothesis extensionally equivalent to the true referent on a restricted
scene pool. Trap T1 targets ZUMU with decoy red \(\wedge\) triangle
(pool: the 24/32 scenes on which decoy and truth agree); trap T2 targets
TIN with the atomic decoy motion = still, perfectly anti-correlated with
parity on a 16-scene pool. Four learning arms face each trap at equal
exchange budget (60), 50 seeds each: (A) naive ostension --- atomic
co-occurrence maximization on random pool scenes; (B) pure statistical
learning --- full-space elimination (2) on random pool scenes, no
minimal pairs, no probes, Occam choice at budget exhaustion; (C-weak)
the full protocol with poisoned ostension but free constructed scenes;
(C-strong) the full protocol with the entire world restricted to the
pool, i.e.~the discriminating scenes \emph{do not exist} --- pure Quine.

\textbf{Table 3.} Injected traps, 50 seeds per arm per trap. Outcomes
per Eq. (6); detection = mean exchange at which the decoy was eliminated
(runs detected/total).

\begin{longtable}[]{@{}lllllll@{}}
\toprule\noalign{}
Trap & Arm & Correct & Kangaroo & Underdet. & Unres. & Detection \\
\midrule\noalign{}
\endhead
\bottomrule\noalign{}
\endlastfoot
T1 (ZUMU) & A naive & 0 & \textbf{50} & 0 & 0 & never \\
T1 & B statistical & 0 & \textbf{50} & 0 & 0 & never \\
T1 & C full, weak & 28 & \textbf{0} & 0 & 22 & 40.6 (50/50) \\
T1 & C full, strong & 0 & \textbf{0} & 9 & 41 & impossible \\
T2 (TIN) & A naive & 0 & \textbf{50} & 0 & 0 & never \\
T2 & B statistical & 0 & \textbf{50} & 0 & 0 & never \\
T2 & C full, weak & 39 & \textbf{0} & 11 & 0 & 13.4 (50/50) \\
T2 & C full, strong & 0 & \textbf{0} & 0 & 50 & impossible \\
\end{longtable}

\begin{figure}[t]\centering
\includegraphics[width=0.95\linewidth]{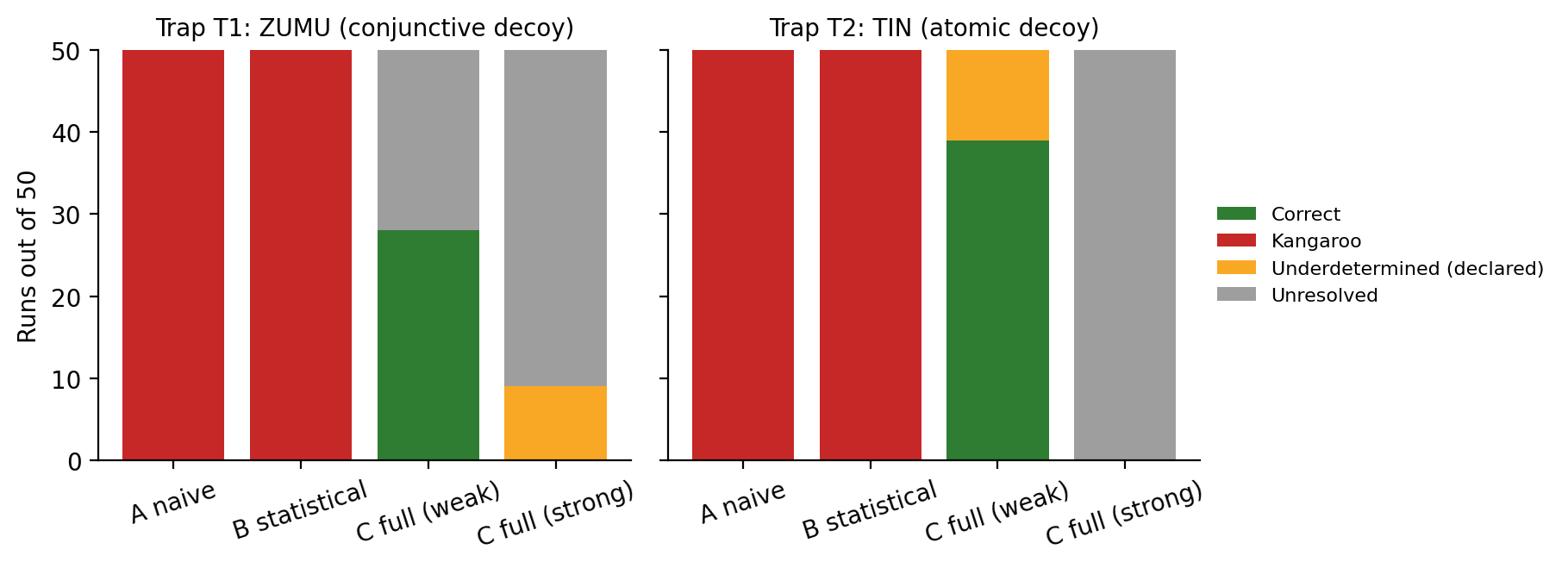}
\caption{Outcome composition under the two injected traps. The passive arms fall into the trap in 100\% of runs; the full protocol never promotes a kangaroo, intercepting the decoy in every weak-trap run and degrading to declared underdetermination or abstention in the strong trap.}\label{fig:3}
\end{figure}

\begin{figure}[t]\centering
\includegraphics[width=0.75\linewidth]{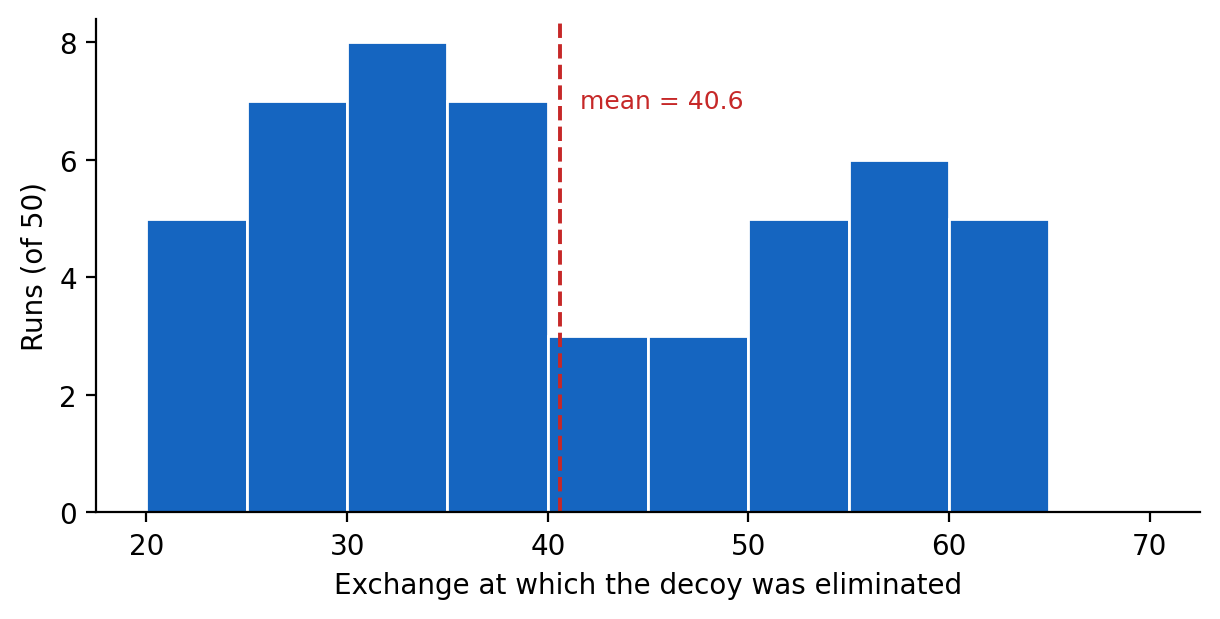}
\caption{Distribution of detection times (trap T1, arm C-weak, 50 seeds): the exchange at which the decoy hypothesis was eliminated. Detection occurred in all 50 runs.}\label{fig:4}
\end{figure}

The contrast is maximal. Arm B is the instructive failure: its
elimination machinery works perfectly, but on the poisoned pool truth
and decoy are indistinguishable, and the deterministic Occam tie-break
lands on the decoy --- a 100\% kangaroo rate produced by a
\emph{correct} statistical learner facing the wrong evidence
distribution. The full protocol promoted no kangaroo in any of the 200
trap runs. In the weak traps it intercepted the decoy in every run
tested, with detection time emerging as a useful new metric: the atomic
decoy (T2) breaks early, at the minimal-pairs stage (mean exchange
13.4), while the conjunctive decoy (T1) survives until probing (mean
40.6, Figure 4).

The strong traps operationalize the extensional form of Quinean
underdetermination. When the discriminating scenes are removed from the
world, no protocol can separate truth from decoy, and the honest output
is not a guess but the \emph{equivalence class}. The class-aware metric
(6) makes this a first-class outcome: in T1-strong, ZUMU is promoted in
9/50 runs with the declared class \{red \(\wedge\) fast, red \(\wedge\)
triangle\} --- counted as kangaroos by a naive metric, but in fact the
strongest claim the evidence supports. T2-strong exhibits the second
degree of caution: in its pool no discriminating minimal pair exists at
all, the contrastive precondition of (5) is never met, and the protocol
abstains entirely (50/50 unresolved). Two distinct fail-safe behaviours
--- declared underdetermination and abstention --- both with zero false
promotion. The non-interference check confirms that the trap does not
damage the rest of the dictionary: across the 50 C-weak runs of T1, the
five non-trapped words yield 178 correct, 58 unresolved, 14
underdetermined, and 0 kangaroos.

\hypertarget{hard-words-and-the-proposer-ablation-does-one-need-an-llm}{%
\subsubsection{5.3 Hard words and the proposer ablation: does one need
an
LLM?}\label{hard-words-and-the-proposer-ablation-does-one-need-an-llm}}

Everything above is achieved by a deterministic script; the LLMs have
served only as informants. Experiment 4 is not a change of topic but the
same architecture applied to a new source of unreliability: where
Sections 5.1--5.2 intercept wrong \emph{referents}, this section
intercepts wrong \emph{proposed rules}, asking where an LLM becomes
necessary on the learner's side and showing that the
intercept-don't-prevent principle keeps it harmless there too. The world
becomes sequential (episodes of six scenes) and the alien lexicon is
extended with two words that lie provably outside the base hypothesis
space (1): \textbf{NUR}, a history-dependent relational word (emitted
iff the current scene has more objects than the previous one), and
\textbf{GUMO}, a contextual homonym with XOR structure (emitted iff
triangle \(\wedge\) odd, or circle \(\wedge\) even) --- a disjunction,
not a conjunction. Outcomes are computed by \emph{extensional}
comparison over all \(33\times 32\) (previous, current) scene pairs:

\[\sigma(h)=\bigl\{(s',s) : h(s',s)\bigr\},\qquad \text{correct} \iff \sigma(\hat h)=\sigma(r_{w}). \qquad (7)\]

Four arms, equal pipeline: \textbf{base} --- the sequential learner over
\(\mathcal{H}_{0}\); \textbf{oracle} --- a hand-extended space
(relational atoms over \(n\), change/same atoms, and all 496 two-clause
disjunctions; 537 hypotheses), representing the ceiling reachable
\emph{if a human already knows which extensions matter}; \textbf{LLM
proposer} --- the base space plus a generate-and-test loop (Figure 6):
when a word's hypothesis set empties, the observation log is shown to an
LLM that proposes up to five candidate rules in a restricted DSL
(conjunctions, relational comparisons on \(n\), attribute-change
predicates, two-clause disjunctions); the script compiles the
candidates, validates them semantically, filters them against the
\emph{entire} log, and subjects survivors to \(M=4\) fresh
pre-registered probes (with the positive-evidence condition) before any
promotion. The LLM proposes; the script disposes. A \textbf{stub} arm
with fixed mixed right/wrong candidates verifies the plumbing only.

\begin{figure}[t]\centering
\includegraphics[width=0.95\linewidth]{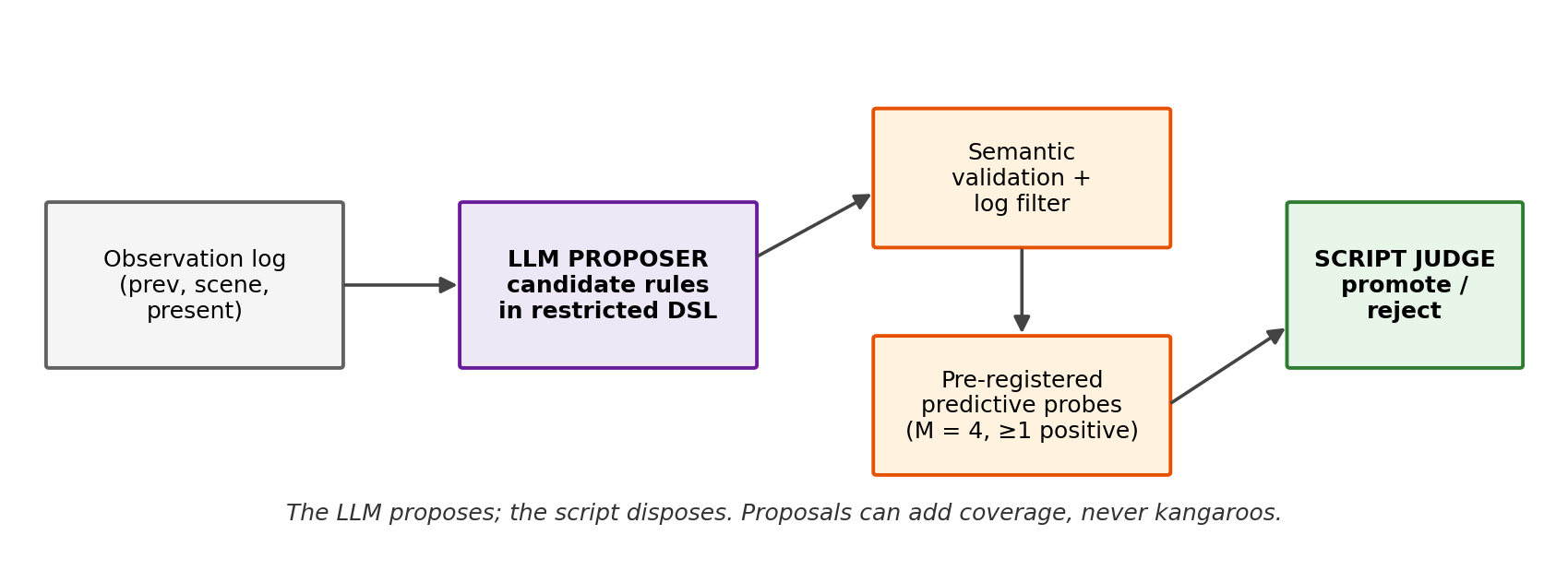}
\caption{The generate-and-test architecture of the proposer arm. Judgment --- semantic validation, retrospective log filter, pre-registered predictive probes --- remains entirely on the script side.}\label{fig:6}
\end{figure}

\textbf{Table 4.} Hard-word outcomes, 30 seeds per arm (protocol version
v3; see ``instrument lessons'' below). Base words (6 per run) are solved
in 100\% of runs in every arm. Precision = log-coherent candidates /
compiled candidates.

\begin{longtable}[]{@{}
  >{\raggedright\arraybackslash}p{(\columnwidth - 12\tabcolsep) * \real{0.1429}}
  >{\raggedright\arraybackslash}p{(\columnwidth - 12\tabcolsep) * \real{0.1429}}
  >{\raggedright\arraybackslash}p{(\columnwidth - 12\tabcolsep) * \real{0.1429}}
  >{\raggedright\arraybackslash}p{(\columnwidth - 12\tabcolsep) * \real{0.1429}}
  >{\raggedright\arraybackslash}p{(\columnwidth - 12\tabcolsep) * \real{0.1429}}
  >{\raggedright\arraybackslash}p{(\columnwidth - 12\tabcolsep) * \real{0.1429}}
  >{\raggedright\arraybackslash}p{(\columnwidth - 12\tabcolsep) * \real{0.1429}}@{}}
\toprule\noalign{}
\begin{minipage}[b]{\linewidth}\raggedright
Proposer
\end{minipage} & \begin{minipage}[b]{\linewidth}\raggedright
NUR correct
\end{minipage} & \begin{minipage}[b]{\linewidth}\raggedright
GUMO correct
\end{minipage} & \begin{minipage}[b]{\linewidth}\raggedright
Hard total
\end{minipage} & \begin{minipage}[b]{\linewidth}\raggedright
Proposals (compiled)
\end{minipage} & \begin{minipage}[b]{\linewidth}\raggedright
Precision
\end{minipage} & \begin{minipage}[b]{\linewidth}\raggedright
Kangaroos
\end{minipage} \\
\midrule\noalign{}
\endhead
\bottomrule\noalign{}
\endlastfoot
None (base space) & 0/30 & 0/30 & 0\% & --- & --- & \textbf{0} \\
Claude Haiku 4.5 & 10/30 & 1/30 & 18\% & 98 & \(\approx\) 11\% &
\textbf{0} \\
Claude Sonnet 4.6 & 28/30 & 15/30 & 72\% & 30 & \(\approx\) 67\% &
\textbf{0} \\
Oracle (hand-coded) & 30/30 & 30/30 & 100\% & --- & --- & \textbf{0} \\
\end{longtable}

\begin{figure}[t]\centering
\includegraphics[width=0.75\linewidth]{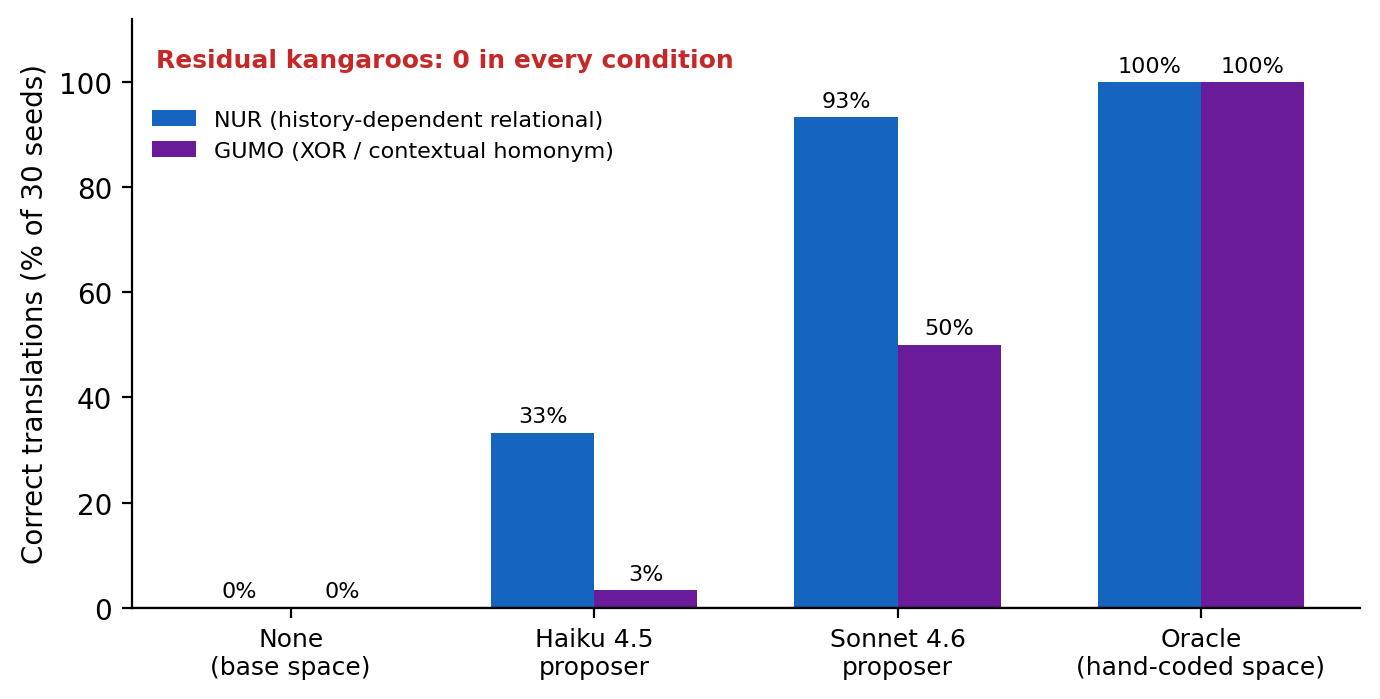}
\caption{The capability ladder on hard words (30 seeds). Coverage scales with proposer capability; the rate of undetected mistranslation is pinned at zero across the entire ladder by the script-side judge.}\label{fig:5}
\end{figure}

Three results. \textbf{(i) Fail-safe degradation.} The base learner
leaves NUR and GUMO unresolved in 30/30 runs each, with zero kangaroos:
a hypothesis space that cannot represent a referent abstains rather than
confabulates. \textbf{(ii) The capability ladder.} Coverage on the hard
words rises monotonically with proposer capability --- 0\%
\(\rightarrow\) 18\% \(\rightarrow\) 72\% \(\rightarrow\) 100\% --- and
the two proposers exhibit distinct styles: Haiku sprays many candidates
of which \(\approx\) 89\% are incoherent with the log; Sonnet proposes
few and targeted (67\% coherent), discovering unaided both the
relational rule and the XOR disjunction in most runs, and reaching 90+\%
of the oracle ceiling on NUR. The per-structure asymmetry (NUR 93\% vs
GUMO 50\% for Sonnet) reproduces, on a new instrument, the classic
concept-learning hierarchy in which disjunctive/XOR structure is hardest
to induce. \textbf{(iii) Capability/safety separation.} Every proposal
that survived the log filter also passed the fresh probes; every
hard-word failure was a failure to \emph{produce} a coherent candidate,
never a wrong candidate promoted; kangaroos are zero in all 120
hard-word runs across all four capability levels, all prompt versions,
and both proposer models. The architecture renders the proposer's
unreliability harmless by construction: proposals can only add coverage,
never kangaroos.

\textbf{Instrument lessons: extended protocol parity.} The proposer arm
taught a methodological lesson worth reporting as a finding. Across
three prompt/parser versions, Sonnet's hard-word performance moved from
30\% to 0\% to 90\% (NUR) --- with the model unchanged. Version 1's
parser lost responses; an added ``abstention'' sentence in version 2
suppressed proposals; raw-response logging then revealed that Sonnet
\emph{reasons in prose before answering} and was being truncated at
\texttt{max\_tokens} = 600 before emitting any JSON, and on one occasion
spontaneously \emph{invented a DSL extension} (an attribute-change atom
inside a disjunction), crashing a validator that checked syntax but not
semantics. Version 3 --- reasoning permitted, JSON required as the final
line, 2000 tokens, a last-valid-list parser, and semantic validation of
every atom --- recovered the model's true capability. The proposer is a
\emph{measuring instrument} whose version comprises code + prompt +
parser + validator; comparisons are valid only at parity of all four,
exactly as numerical comparisons are valid only at parity of code
version. We label all runs accordingly and archive raw proposer
responses alongside the results, because for the LLM arm --- the only
arm whose re-execution is not guaranteed to reproduce (the model behind
an API name can change) --- the raw log is the sole ground truth.

\hypertarget{noisy-informants-the-validity-domain-of-the-zero-kangaroo-result}{%
\subsubsection{5.4 Noisy informants: the validity domain of the
zero-kangaroo
result}\label{noisy-informants-the-validity-domain-of-the-zero-kangaroo-result}}

All results above assume a perfectly faithful speaker. Experiment 5
removes that assumption: a noisy SIM speaker flips the presence of each
L2 word independently with probability \(p\) per utterance (symmetric
channel noise, dedicated seeded RNG, fully deterministic), and the full
protocol (active + R2) runs unchanged, 50 seeds per noise level. The
pre-registered question was honest by construction: elimination (2)
assumes fidelity, and under noise it \emph{can} delete the true
referent, so either the quarantine machinery contains the damage or
kangaroos appear and delimit the validity domain.

\textbf{Table 5.} Robustness under per-word flip noise (active + R2, 50
seeds per level). Fidelity = utterance-level agreement with the lexicon.

\begin{longtable}[]{@{}
  >{\raggedright\arraybackslash}p{(\columnwidth - 10\tabcolsep) * \real{0.1667}}
  >{\raggedright\arraybackslash}p{(\columnwidth - 10\tabcolsep) * \real{0.1667}}
  >{\raggedright\arraybackslash}p{(\columnwidth - 10\tabcolsep) * \real{0.1667}}
  >{\raggedright\arraybackslash}p{(\columnwidth - 10\tabcolsep) * \real{0.1667}}
  >{\raggedright\arraybackslash}p{(\columnwidth - 10\tabcolsep) * \real{0.1667}}
  >{\raggedright\arraybackslash}p{(\columnwidth - 10\tabcolsep) * \real{0.1667}}@{}}
\toprule\noalign{}
\begin{minipage}[b]{\linewidth}\raggedright
Noise \(p\)
\end{minipage} & \begin{minipage}[b]{\linewidth}\raggedright
Fidelity
\end{minipage} & \begin{minipage}[b]{\linewidth}\raggedright
Correct
\end{minipage} & \begin{minipage}[b]{\linewidth}\raggedright
Kangaroos
\end{minipage} & \begin{minipage}[b]{\linewidth}\raggedright
Unresolved
\end{minipage} & \begin{minipage}[b]{\linewidth}\raggedright
Total kangaroos
\end{minipage} \\
\midrule\noalign{}
\endhead
\bottomrule\noalign{}
\endlastfoot
0\% & 100.0\% & 4.72 ± 0.81 & \textbf{0.00 ± 0.00} & 1.28 ± 0.81 & 0 \\
2\% & 88.6\% & 3.54 ± 1.11 & \textbf{0.00 ± 0.00} & 2.46 ± 1.11 & 0 \\
5\% & 74.2\% & 2.38 ± 1.03 & 0.08 ± 0.34 & 3.54 ± 0.97 & 4 \\
10\% & 53.1\% & 1.16 ± 0.91 & 0.10 ± 0.30 & 4.74 ± 0.92 & 5 \\
\end{longtable}

\begin{figure}[t]\centering
\includegraphics[width=0.75\linewidth]{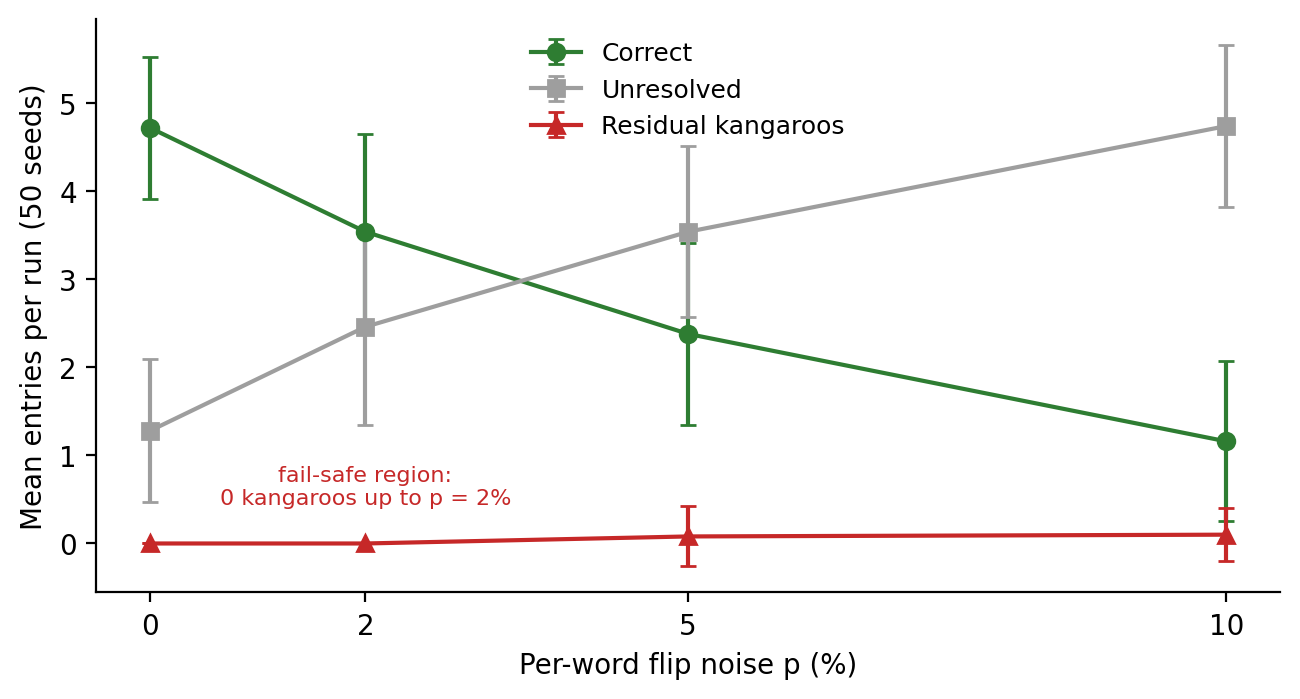}
\caption{Graceful degradation under informant noise. Coverage collapses far faster than correctness: at 10\% per-word noise (utterance fidelity 53\%) the protocol resolves little, but errs almost never --- degradation is by abstention.}\label{fig:7}
\end{figure}

The result delimits and, within its limits, strengthens the claim. The
zero-kangaroo guarantee survives intact at 2\% per-word noise --- i.e.,
with more than one utterance in ten corrupted --- with the entire cost
paid in coverage. At 5--10\% a small leak appears (4--5 kangaroos across
100 runs): a noisy observation occasionally deletes the true referent
and a surviving rival then passes its probes by chance. Even there the
failure profile is the designed one: at \(p = 10\%\) the protocol leaves
4.74 entries per run unresolved and mistranslates 0.10 --- it
predominantly abstains rather than errs. Extending elimination to a
probabilistic scoring rule in the style of Fazly et al.~(2010), with the
same pre-registration and quarantine wrapper, is the natural next step
for high-noise regimes.

\hypertarget{discussion}{%
\subsection{6. Discussion}\label{discussion}}

\textbf{Correctness is the protocol's; coverage is the instruments'.}
The headline of Table 2 is not 4.72 correct entries but the zero beside
it; the headline of Table 4 is not Sonnet's 72\% but the zero that holds
from no-proposer to oracle. Across every experiment, every adversarial
manipulation, and every instrument version in this paper --- 200+ runs,
two informant species, two proposer models, three prompt versions ---
the number of promoted-and-wrong translations that survived to the end
of a run under the full protocol is zero at speaker fidelity \(\ge\)
88.6\%, and at most 0.10 per run at fidelity as low as 53\%, while one
mechanism after another (recovery round, traps, proposers, noise) moved
coverage up and down by whole units. This decoupling is the paper's
central claim, and it is an architectural one: pre-registered
prediction, actively constructed disagreement, contrastive
preconditions, quarantine, and script-side judgment are jointly designed
so that, within the tested regime, the system's confidence fails only
\emph{downward} --- into abstention or declared underdetermination ---
rather than sideways into silent error; Section 5.4 measures where that
regime ends.

\textbf{One form of Quinean underdetermination, operationalized twice.}
We do not claim to capture indeterminacy of translation in general ---
only its extensional form within a finite constructed world. Within that
scope, the kangaroo effect makes it measurable (a rate, a detection
time); the strong traps make it irreducible by design; and the
class-aware metric (6) shows what a rational agent should do there:
report the equivalence class. The two degrees of caution observed ---
declared underdetermination when contrastive evidence exists within the
pool (T1-strong), full abstention when it does not (T2-strong) ---
provide, to our knowledge, the first experimental separation of
underdetermination into evidentially distinct grades in a setting of
this kind.

\textbf{The two roles of the LLM, and the symmetry between them.} In
Sections 3--5.2 the LLM is the \emph{specimen}: a synthetic native
informant whose interpretive behaviour under an imposed lexicon
(presupposition of plurality, per-object doubling --- deterministic
readings of underspecified rules) is itself data. In Section 5.3 it
becomes the \emph{instrument}: a hypothesis generator for spaces no one
hand-coded. In both roles the same discipline applies --- the judgment
never leaves the script --- and in both the same philosophy holds: the
kangaroo is not prevented but intercepted, whether the kangaroo is a
wrong referent or a wrong proposed rule. That an architecture designed
against Quinean underdetermination turns out, unchanged, to be an
architecture for safely harnessing unreliable generative models is, we
believe, the most transferable result of this work. In alignment terms,
the protocol is a behavioural concept-identity check: it certifies, by
survival under active falsification, that informant and learner are
attached to the same referent --- the property that Level B will need to
certify for latent-space maps.

\textbf{Relation to XSL, completed.} Our base learner is Siskind's
(1996) eliminative XSL with three additions that his analysis motivates:
verification before commitment (pre-registered probes where his
algorithm commits on convergence), active evidence construction (his
learner is passive), and first-class revocation (his corrupted entries
can only be superseded, not repaired). The trap experiment then
quantifies precisely the gap these additions close: eliminative learning
without them falls into an injected confound in 100\% of runs.

\hypertarget{limitations-and-future-work}{%
\subsection{7. Limitations and future
work}\label{limitations-and-future-work}}

The micro-world is small (32 scenes) and, although channel noise is now
tested up to 10\% per word (Section 5.4), Siskind-scale referential
uncertainty, synonymy, polysemy (as distinct from our contextual
homonymy), graded or probabilistic referents, and hierarchical category
systems remain untested; scaling the ontology is the most direct way to
address the legitimate concern that small worlds flatter eliminative
learners. The recovery round is single-shot: trap-T1-type residuals
suggest a third round would convert remaining near-misses, and the round
count is a natural knob for a coverage/cost frontier. The proposer DSL
is intentionally narrow; widening it shifts the burden to semantic
validation, and the observed tendency of stronger models to generalize
beyond the contract suggests validation must scale with proposer
capability. The LLM informant is stationary by construction (temperature
0, fixed prompt); a deliberately \emph{non-stationary} or deceptive
informant is the natural adversarial sequel to the trap experiment.
Finally, Level B of the programme --- alignment of latent representation
spaces between open-weight models via Procrustes maps over anchor sets,
with a ``ladder of games'' for evaluative concepts --- is where the
astrolinguistic question becomes geometric; the protocol of this paper
is designed to serve there as the behavioural verification layer for
claimed alignments.

\hypertarget{data-availability}{%
\subsection{Data availability}\label{data-availability}}

All code (orchestrator, campaign runner, trap and hard-condition
suites), per-seed CSV results, JSONL interaction logs, and raw proposer
responses are archived with the version labels described in Section 5.3
and are available from the authors; SIM results are bit-reproducible
from seeds. Total API cost of all reported LLM campaigns: \textless{}
EUR 5. The noise experiment (Section 5.4) is SIM-only and free.

\hypertarget{acknowledgements}{%
\subsection{Acknowledgements}\label{acknowledgements}}

The authors thank the colleagues who commented on earlier drafts. The
large language models used in the experiments, all accessed through
their public APIs at temperature 0 with fixed prompts, are documented in
the text: OpenAI GPT-4o-mini served as the L2 informant; Anthropic
Claude Haiku 4.5 served as the L1 informant and as a hypothesis
proposer; Anthropic Claude Sonnet 4.6 served as a hypothesis proposer
(Section 5.3). Anthropic Claude was additionally used as a coding
assistant for the experimental scripts, under the conventions of Section
4. The authors take full responsibility for the content of this
manuscript.

\hypertarget{references}{%
\subsection{References}\label{references}}

Christiano, P., Cotra, A. and Xu, M. (2021). `Eliciting latent
knowledge: how to tell if your eyes deceive you', Alignment Research
Center, technical report.

Conneau, A., Lample, G., Ranzato, M., Denoyer, L. and Jégou, H. (2017).
`Word translation without parallel data', \emph{arXiv preprint}
arXiv:1710.04087.

Euzenat, J. and Shvaiko, P. (2013). \emph{Ontology Matching}, 2nd edn.
Berlin: Springer.

Fazly, A., Alishahi, A. and Stevenson, S. (2010). `A probabilistic
computational model of cross-situational word learning', \emph{Cognitive
Science}, 34(6): 1017--1063.

Freudenthal, H. (1960). \emph{Lincos: Design of a Language for Cosmic
Intercourse, Part I}. Amsterdam: North-Holland.

Galantucci, B. (2005). `An experimental study of the emergence of human
communication systems', \emph{Cognitive Science}, 29(5): 737--767.

Irving, G., Christiano, P. and Amodei, D. (2018). `AI safety via
debate', \emph{arXiv preprint} arXiv:1805.00899.

Kouwenhoven, T., Peeperkorn, M. and Verhoef, T. (2024). `Searching for
structure: investigating emergent communication with large language
models', \emph{arXiv preprint} arXiv:2412.07646.

Lazaridou, A., Peysakhovich, A. and Baroni, M. (2016). `Multi-agent
cooperation and the emergence of (natural) language', \emph{arXiv
preprint} arXiv:1612.07182; see also Lazaridou, A. et al.~(2016)
`Towards multi-agent communication-based language learning',
arXiv:1605.07133.

Maiorca, V., Moschella, L., Norelli, A., Fumero, M., Locatello, F. and
Rodolà, E. (2023). `Latent space translation via semantic alignment',
\emph{Advances in Neural Information Processing Systems}, 36.

Moschella, L., Maiorca, V., Fumero, M., Norelli, A., Locatello, F. and
Rodolà, E. (2023). `Relative representations enable zero-shot latent
space communication', \emph{Proceedings of ICLR 2023}.

Ollongren, A. (2013). \emph{Astrolinguistics: Design of a Linguistic
System for Interstellar Communication Based on Logic}. New York:
Springer.

Quine, W. V. O. (1960). \emph{Word and Object}. Cambridge, MA: MIT
Press.

Siskind, J. M. (1996). `A computational study of cross-situational
techniques for learning word-to-meaning mappings', \emph{Cognition},
61(1--2): 39--91.

Smith, K., Smith, A. D. M. and Blythe, R. A. (2011). `Cross-situational
learning: an experimental study of word-learning mechanisms',
\emph{Cognitive Science}, 35(3): 480--498.

Steels, L. (2015). \emph{The Talking Heads Experiment: Origins of Words
and Meanings}. Berlin: Language Science Press.

Sucholutsky, I., Muttenthaler, L., Weller, A., et al.~(2023). `Getting
aligned on representational alignment', \emph{arXiv preprint}
arXiv:2310.13018.

\end{document}